\documentclass[10pt,twocolumn,letterpaper]{article}

\usepackage[pagenumbers]{cvpr} % To force page numbers, e.g. for an arXiv version

\definecolor{cvprblue}{rgb}{0.21,0.49,0.74}
\usepackage[pagebackref,breaklinks,colorlinks,allcolors=cvprblue]{hyperref}

\usepackage{bm}
\usepackage{amsmath}
\usepackage{tabularx}
\newcolumntype{C}{>{\centering\arraybackslash}X}
\usepackage{wrapfig}

\usepackage{ragged2e}
\usepackage{xspace}
\usepackage{makecell} 

\title{Style3D: Attention-guided Multi-view Style Transfer for 3D Object Generation}

\author{Bingjie Song, Xin Huang, Ruting Xie, Xue Wang, Qing Wang\\ %{$^\dag$}
Northwestern Polytechnical University\\
}

\begin{document}
\twocolumn[{%
\renewcommand\twocolumn[1][]{#1}%
\maketitle
\begin{center}
    \centering
    \vspace{-4mm}
    \captionsetup{type=figure}
    \includegraphics[width=1\hsize]{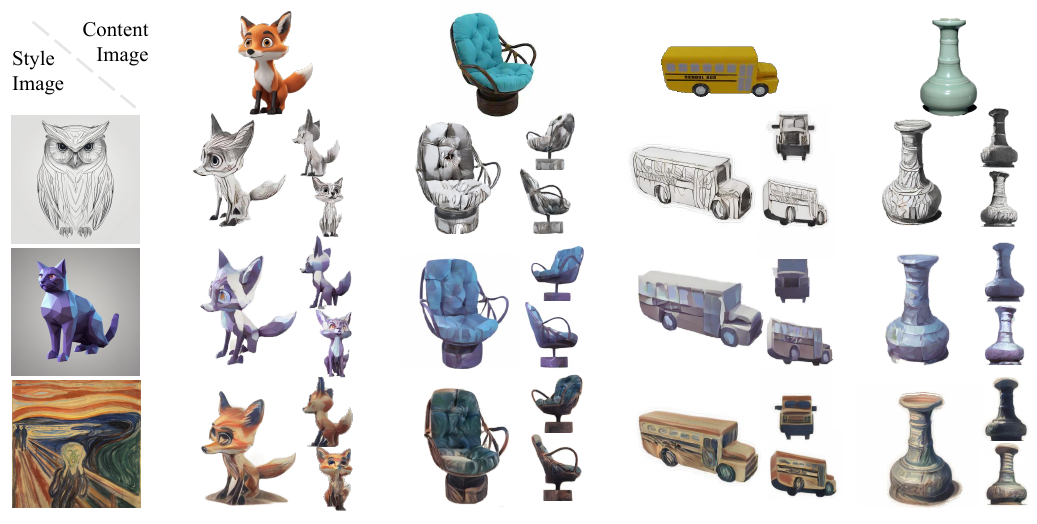}
    \vspace{-2mm}
    \caption{\textbf{Style3D:} a novel approach for instant generation of stylized 3D objects. It generates high-quality, stylized 3D objects from single content-style image pairs in 30 seconds, preserving geometric consistency and stylistic fidelity across multiple viewpoints.
    }  
    % \caption{\textbf{Style3D.} Visual results of Style3D, showcasing the stylized 3D objects generated from different content images and style images, demonstrating how our method maintains geometric consistency and stylistic fidelity across multiple perspectives.
    % }  
    \label{fig:teaser}
\end{center}%
}]

\begin{abstract}
We present Style3D, a novel approach for generating stylized 3D objects from a content image and a style image. Unlike most previous methods that require case- or style-specific training, Style3D supports instant 3D object stylization. Our key insight is that 3D object stylization can be decomposed into two interconnected processes: multi-view dual-feature alignment and sparse-view spatial reconstruction. We introduce MultiFusion Attention, an attention-guided technique to achieve multi-view stylization from the content-style pair. Specifically, the query features from the content image preserve geometric consistency across multiple views, while the key and value features from the style image are used to guide the stylistic transfer. This dual-feature alignment ensures that spatial coherence and stylistic fidelity are maintained across multi-view images. Finally, a large 3D reconstruction model is introduced to generate coherent stylized 3D objects. By establishing an interplay between structural and stylistic features across multiple views, our approach enables a holistic 3D stylization process. Extensive experiments demonstrate that Style3D offers a more flexible and scalable solution for generating style-consistent 3D assets, surpassing existing methods in both computational efficiency and visual quality.

\end{abstract}    
\section{Introduction}
\label{sec:intro}

Recent advancements in 3D generation~\cite{poole2023dreamfusion,wang2023prolificdreamer,sun2023dreamcraft3d} have significantly lowered the barrier to creating 3D assets, unlocking new possibilities for applications in video games, digital art, virtual reality, and beyond. Despite these advances in generating individual 3D models, creating 3D objects with arbitrary, user-defined styles remains a key challenge. Most existing methods focus on generating 3D objects from a text prompt or a single image, making it difficult to transfer diverse styles onto these objects. However, enabling users to generate 3D objects with flexible styling is essential for creating unique assets that fit a wide range of creative and interactive contexts.

% Recent advancements in 3D generation have reduced the barrier to creating 3D assets, opening new possibilities for diverse applications such as gaming, digital art, and virtual reality. However, despite the advances in generating individual 3D models, one of the obstacles remains unresolved: the challenge of creating multiple, style-consistent 3D objects with [[[high geometry fidelity and minimal input -- without the need for 3D meshes or specific retrainting, especially when mesh inputs are unavailable or impractical.]]][[where rapid and consistent generation of 3D assets is a prerequisite.]]The key [challenge] lies not just in the [visual aesthetic] but also in the spatial coherence between the generated 3D structure and its [stylistic transformation] across different views.
While style transfer techniques have been successfully applied in 2D domains~\cite{kwon2022clipstylerimagestyletransfer,wang2023stylediffusioncontrollabledisentangledstyle,wang2023stylediffusioncontrollabledisentangledstyle}, few methods have extended this capability to broad 3D object generation. Some methods~\cite{zhang2022arfartisticradiancefields,liu2023stylerfzeroshot3dstyle} attempt to achieve 3D style transfer by applying 2D style techniques to dense multi-view images and then reconstructing 3D contents. Others~\cite{zhuang2023dreameditortextdriven3dscene,chen2023text2textextdriventexturesynthesis,metzer2022latentnerf} focus on editing or modifying existing 3D assets with a specific style, rather than creating novel, stylized 3D objects from scratch. In general, these approaches come with notable limitations: (1) Multi-view inconsistency. By applying 2D style transfer independently to each view, these methods often introduce inconsistencies across multi-view images, leading to 3D reconstructions with poor stylization coherence. (2) Time-consuming. These approaches typically require style-specific training, making them slow and inefficient for new styles and new objects due to lengthy optimization processes. Recently, methods like StyleRF~\cite{liu2023stylerfzeroshot3dstyle} have introduced zero-shot 3D style transfer, enabling the application of an arbitrary style to a specific 3D object without the need for style-specific training. However, they require retraining the style transfer module each time a new object is introduced. This retraining process increases computational demands and limits flexibility. Consequently, these issues highlight a critical need for the capability of generating diverse, stylistically-consistent 3D objects directly, without relying on slow, case-by-case optimization.  

In this paper, we propose Style3D, a novel diffusion-based approach that instantly generates 3D stylized objects from content-style image pairs. The core idea of Style3D lies in decomposing 3D object stylization into two interconnected tasks: multi-view dual-feature alignment and sparse-view spatial reconstruction. This framework ensures the preservation of geometric integrity across all views while adopting the visual characteristics of the style image without the need for training. For consistent stylistic multi-view generation, we introduce a MultiFusion Attention mechanism that orchestrates the transfer of stylistic information. Specifically, the query features of the content image anchor the spatial layout and geometric consistency across the multi-view generation, while the key and value features from the style image direct the stylistic transformation. These features are fused through the reverse process to ensure geometric coherence and stylistic fidelity across different views. Furthermore, our high-fidelity 3D reconstruction process integrates these stylistically aligned views into a unified, coherent triplane representation, followed by a network to predict SDF, color and weight. By establishing an interplay between structural and stylistic features, Style3D enables a holistic 3D stylization process that significantly improves the efficiency and scalability of generating large numbers of high-quality stylized 3D objects. To summarize, our core contributions are:
%providing a seamless transition from stylized images to a fully realized 3D object

\begin{enumerate}
    \setlength\itemsep{0pt}
    \item We introduce a novel framework for instant stylized 3D object generation by decomposing the task into multi-view dual-feature alignment and sparse-view spatial reconstruction.
    % \item We introduce a novel framework that decomposes the 3D object stylization task into two feasible sub-tasks: the first stage integrates spatial layout information with stylistic features to the generated multi-view images, while the second stage employs a transformer-based encoder-decoder architecture to reconstruct the stylized 3D object from the multi-view images inheriting these dual features.
    %We introduce a framework, that contains multi-view stylistic alignment and sparse-view spatial reconstruction, for instantly generating stylized 3D objects without case- and style-specific training.
    %在3d object风格化的任务上首次把任务分解为了两个可行的子任务并建立了相应的处理方法，第一阶段将空间布局信息与风格信息集成到多视角，第二阶段应用从多视角图像中继承到的高维特征完成这个特殊的三维重建任务
    \item We propose an attention mechanism that facilitates the fusion of spatial-style features through a multi-view attention fusion process, allowing consistent multi-view stylization. 
    % \item We propose an attention mechanism that facilitates the fusion of spatial domain and style domain features through a multi-view attention fusion process. This mechanism takes spatial domain, style domain, and multi-view control as conditions, while allowing parameters to regulate the balance between the strength of stylization and the preservation of the target content.
    %We propose an attention mechanism to bind the geometric features from the content image with the stylistic features from the style image, ensuring that both geometry and style are preserved and adapted effectively across different views.  
    %注意力视为对目标的空间域特征和风格图的风格域特征的映射体现？并设置了一种多视角注意力融合的机制，把空间域、风格域、多视角控制作为条件，并允许参数控制风格化强度与目标内容的保持程度
    %这个是从一个2d方法中借鉴来的，迁移到了6视图生成的任务上，但是操作没有什么太本质的区别，就是把注意力中的内容图的Q和风格图的KV结合起来再去做一次生成。所以为了和他们区分开并找出一些创新，没有非常明确的写出各种操作手段，而是建模为了一个注意力处理器MultiFusion Attention，适配于我们一阶段的6视图风格化任务
    \item We integrate a robust 3D reconstruction process based on a triplane representation, ensuring geometrically accurate and stylistically consistent results.
    % We integrate a robust 3D reconstruction process that encodes features from the multi-view images, and then decodes them to construct a triplane representation with subsequent optimization. This encoding-decoding approach effectively preserves the spatial and stylistic information necessary for our task in the second stage, maintaining strong consistency with the objective.
    % We integrate a robust 3D reconstruction process that assembles the stylized multi-view images into a coherent 3D object, ensuring geometrically accurate and stylistically consistent results.
    %这个确实没有特别大的创新，基本就是把现行的大家在（不是3d风格化的）普通3d重建或3d生成重建阶段的常用手段复用了一遍，大家的方法都大差不差，先进行特征编码、然后解码并建立三平面表示、然后使用flexicubes获得mesh并进行一些优化。可能只能说这种路线与我们的任务具有较好的一致性，例如这种编解码的方式很好地从二阶段输入的多视图中保持了我们任务所需要的空间和风格信息，并具有较快的重建速度
\end{enumerate}

\section{Related Work}
\label{sec:relatedwork}

\begin{figure*}[t]
\includegraphics[width=0.95\textwidth]{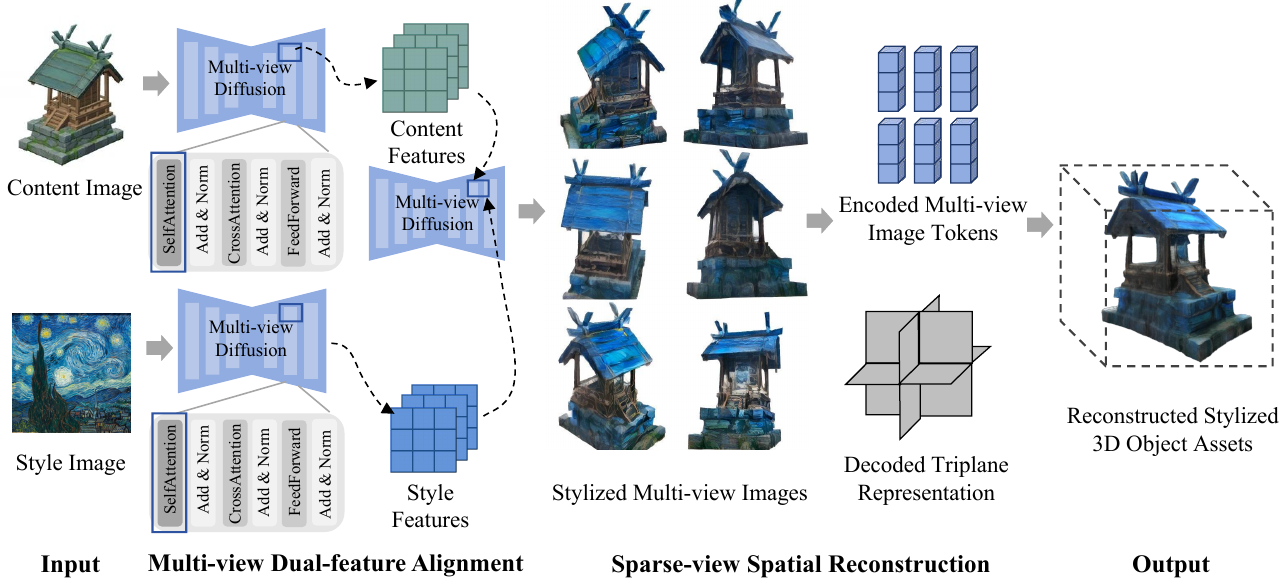}
\vspace{-3pt}
\caption{\textbf{Method overview.} Given two input images, one serving as target content and the other as target style, we first perform Multi-view Dual-feature Alignment in the first stage. This involves extracting content features and style features in the multi-view diffusion process, which separately anchor the geometric and stylistic characteristics. Then these features are fused using an attention mechanism to generate multiple stylized views of the object. In the second stage, we leverage Sparse-view Spatial Reconstruction, where the generated multi-view images are passed through a feature encoder and decoded into a 3D object. The decoder works by integrating spatial and stylistic features across a triplane presentation to produce a coherent 3D mesh. The entire process seamlessly integrates style and geometry while maintaining high computational efficiency, resulting in a stylized 3D object as the final output. }
\vspace{-7pt}
\label{fig:overview}
\end{figure*}

\noindent\textbf{3D Object Generation.} Early works in 3D object generation focus on leveraging 2D diffusion models to produce 3D assets, rather than directly training 3D diffusion models. DreamFusion~\cite{poole2023dreamfusion} first introduces the SDS loss to distill 3D content from a pre-trained image diffusion model. Following this, several studies explore combining SDS loss with various 3D representations and refining the optimization framework~\cite{tang2023make, wang2023prolificdreamer,lin2023magic3d,huang2024humannorm,sun2023dreamcraft3d,wang2024themestation}. However, these per-scene training approaches are inherently time-consuming and computationally expensive. With the advent of large-scale 3D datasets~\cite{deitke2023objaverse}, some approaches have emerged that leverage feed-forward 3D generative models~\cite{nichol2022point,jun2023shap}, which offer improved generalization and enable faster 3D model generation. LRM~\cite{hong2024lrm} demonstrates the effectiveness of transformer-based architectures in mapping image tokens to triplanes for 3D representation. Building on this, Instant3D~\cite{li2024instant3d} extends the LRM framework to incorporate multi-view images, resulting in significant improvements in the quality and accuracy of 3D reconstructions. Other models, such as LGM~\cite{tang2024lgm} and GRM~\cite{xu2024grm}, replace the triplane representation with 3D Gaussian splatting, thereby enhancing rendering efficiency. InstantMesh~\cite{xu2024instantmesh} introduces a mesh-based representation and additional geometric supervision, leading to improvements in both training efficiency and reconstruction quality. However, these approaches generally focus on generating 3D contents from text prompts or single images, and they face challenges in producing results with flexible or diverse styles.
% (e.g., SyncDreamer~\cite{liu2024syncdreamergeneratingmultiviewconsistentimages} Wonder3D~\cite{long2024wonder3d}) use volumetric rendering methods for 3D reconstruction. One-2-3-45++~\cite{liu2023one2345fastsingleimage} first generates multi-view images using Zero-123~\cite{shi2023zero123singleimageconsistent}, then reconstructs 3D geometry.

\noindent\textbf{2D Image Stylization.} 
% Neural Style Transfer (NST)~\cite{7780634} has been a foundational approach in image stylization. However, its iterative optimization process, aligning content images to style images, is computationally intensive. To solve this,
Early works such as CLIPstyler~\cite{kwon2022clipstylerimagestyletransfer} and StyleClip~\cite{patashnik2021stylecliptextdrivenmanipulationstylegan} explore using text input as a condition for style transfer and image generation based on the CLIP model~\cite{radford2021learning}. With advancements in diffusion models~\cite{ho2020denoisingdiffusionprobabilisticmodels}, some approaches~\cite{wang2023stylediffusioncontrollabledisentangledstyle,huang2023diffstylercontrollabledualdiffusion,sohn2023styledroptexttoimagegenerationstyle} leverage the pre-trained diffusion priors to directly generate stylized images. InST~\cite{zhang2023inversionbasedstyletransferdiffusion} introduces a textual inversion-based approach, mapping a given style to corresponding textual embeddings. Diffusion in Style~\cite{Everaert_2023_ICCV} utilizes diffusion models in a more generalized form, injecting various artistic styles into a given image. Recently, Style Injection In Diffusion~\cite{chung2024styleinjectiondiffusiontrainingfree} and RB-Modulation~\cite{rout2024rbmodulationtrainingfreepersonalizationdiffusion} modify attention layers to enhance fidelity and variability, supporting more controlled image generation. StyleAligned~\cite{hertz2024stylealignedimagegeneration} employs shared attention mechanisms to align content and style effectively. By combining diffusion models with techniques like disentangling representations~\cite{wang2023stylediffusioncontrollabledisentangledstyle}, zero-shot learning~\cite{yang2023zeroshotcontrastivelosstextguided,rout2024rbmodulationtrainingfreepersonalizationdiffusion,huang2023diffstylercontrollabledualdiffusion}, and multimodal inputs~\cite{han2024styleboothimagestyleediting}, these methods provide new possibilities for both research and practical applications in image stylization. However, these methods still face challenges in generating consistently stylized multi-view images, limiting their effectiveness in stylized 3D object generation.

\noindent\textbf{3D Stylization.} Recent advancements have extended style transfer to 3D reconstruction and generation. Several methods~\cite{chen2022tangotextdrivenphotorealisticrobust,metzer2022latentnerf,chen2023text2textextdriventexturesynthesis,richardson2023texturetextguidedtexturing3d,zhang2024mapatextdrivenphotorealisticmaterial} focus on generating textures with varying styles for 3D objects, utilizing pre-trained diffusion models to generate textures based on input text prompts or images. For instance, Text2Tex~\cite{chen2023text2textextdriventexturesynthesis} and TEXTure~\cite{richardson2023texturetextguidedtexturing3d} employ diffusion models conditioned on depth maps to progressively generate high-quality textures. However, these methods are limited in generating textures from a style image, as the texture's style is determined by prompts or a content image. Other approaches explore style transfer within 3D representations. ARF~\cite{zhang2022arfartisticradiancefields} introduces style losses into the optimization of NeRF~\cite{mildenhall2020nerfrepresentingscenesneural} to generate art-stylized 3D scenes. UPST-NeRF~\cite{chen2022upstnerfuniversalphotorealisticstyle} proposes a universal photorealistic style transfer approach that preserves the original geometric and lighting characteristics during stylization. StyleRF~\cite{liu2023stylerfzeroshot3dstyle} achieves zero-shot style transfer through a sampling-invariant content transformation. 3DStyleNet~\cite{yin20213dstylenetcreating3dshapes} generates 3D shapes with both geometric and texture style variations. StylizedGS~\cite{zhang2024stylizedgscontrollablestylization3d} enables controllable stylization for 3D Gaussian splatting~\cite{kerbl20233d}. Despite these advancements, instant 3D scene stylization with arbitrary styles remains challenging due to slow, case- or style-specific optimization. In contrast, our feed-forward approach leverages diffusion priors to generate consistent stylized multi-view images, enabling faster 3D object stylization without the need for additional training.

\section{Method}
\label{sec:method}

In this section, we describe the core components of our Style3D framework, as shown in Fig.~\ref{fig:overview}, a novel method for 3D style transfer and generation. The key objective of Style3D is to maintain the geometric consistency of 3D objects while effectively adopting the desired style. To achieve this, we find it an efficient solution to decompose the process into two distinct but interrelated stages: \textit{Multi-View Dual-Feature Alignment} and \textit{Sparse-view Spatial Reconstruction}.
The former stage (Sec.~\ref{sec:stage1}) generates multiple stylized views of the object by aligning the content and style features through an attention-guided mechanism called Multi-Fusion Attention. The latter stage (Sec.~\ref{sec:stage2}) utilizes these generated views for reconstructing the 3D object by encoding the features and applying a transformer-based triplane decoder to generate the final high-quality, stylized object.

 %The former stage leverages an attention-guided approach to control the balance between geometry and style transfer across different views, and the latter is a transformer-based triplane encoder-decoder structure enabling the generation of high-quality, stylized 3D objects.

%Diffusion models, such as Stable Diffusion, generate images by iteratively refining noisy inputs, using a process of denoising guided by learned priors. Zero123++ is a diffusion-based model for multi-view 3D generation, it uses a 3x2 tiling approach, generating six views arranged to capture the joint distribution of multi-view images, improving consistency. Zero123++ also enhances conditioning mechanisms by incorporating multiple strategies to better utilize pre-trained diffusion priors, ensuring stylistic coherence across views. Furthermore, it revises the resolution approach by using a new noise schedule and training process, overcoming the instability issues seen in Zero-1-to-3, which used lower resolution. This enables higher-resolution training, improving global structure preservation and multi-view consistency.

\subsection{Multi-View Dual-Feature Alignment}
\label{sec:stage1}

We draw inspiration from advancements in 2D style transfer~\cite{hertz2024stylealignedimagegeneration,chung2024styleinjectiondiffusiontrainingfree,dathathri2020plugplaylanguagemodels,jeong2024visualstylepromptingswapping}. They have discovered that attention maps and key-value features play critical roles in shaping spatial structure and blending content with style. Additionally, while existing 3D generation frameworks~\cite{shi2023zero123singleimageconsistent,xu2024instantmesh,long2024wonder3d,shi2024mvdreammultiviewdiffusion3d} demonstrate potential, they face limitations in adapting to diverse styles. Our main insight is that consistently aligning style and content features across multiple views, textures, and colors from the style images can be effectively integrated with the geometric structure of the content images, paving the way for 3D generation. To achieve this, we employ a Multi-view Dual-feature Alignment process with advanced attention mechanisms.

\noindent\textbf{Style Transfer in Multi-View}. In our approach, we adapt the pre-trained single-image-to-multi-view model, Zero123++~\cite{shi2023zero123singleimageconsistent}, as the backbone and enhance its capacity by modifying the attention layers. Zero123++ arranges six views in a 3$\times$2 tiling format to tackle the challenge of maintaining consistency in different views by modeling their joint distribution. Additionally, Zero123++ employs an improved noise schedule and optimized resolution to enhance stability and preserve global structural integrity. 
% \noindent\textbf{Style Transfer in Multi-View}. In our approach, we leverage the power of the robust, pre-trained multi-view diffusion framework Zero123++~\cite{} with our critical modifications to the attention layers. Zero123++ arranges six views in a $3\times2$ tiling format to tackle the challenge of maintaining consistency in different views by modeling their joint distribution, and also employs an improved noise schedule and optimized resolution to enhance stability and preserve global structural integrity. 

We omit the specific description of the classical diffusion process~\cite{rombach2021highresolution} with attention computation and instead mainly focus on introducing our approach. Specifically, we perform multi-view DDPM~\cite{ho2020denoisingdiffusionprobabilisticmodels} inversion on both the content and style images to extract their latent distributions, adopting a dual U-Net architecture to create an advanced conditioning framework (see Fig.~\ref{fig:overview}, left). The primary U-Net captures essential spatial and geometric relationships across multiple views, stabilizing core structural features, while the auxiliary conditioning U-Net infuses local, context-sensitive style attributes. This hierarchical configuration allows nuanced control over the balance between style and content, thereby enhancing the fidelity of multi-view 3D generation.
% We omit the specific description of the classical diffusion process~\cite{} with attention computation, and instead mainly focus on introducing our approach. We perform multi-view DDPM~\cite{} inversion on both the content and style images, as shown in the left part of Fig.~\ref{fig:overview}, to extract their latent distributions and adopt a dual U-Net architecture, as in Zero123++, to establish an advanced conditioning framework. The primary U-Net captures overarching spatial and geometric relationships across multiple views, stabilizing core structural features, while the auxiliary conditioning U-Net orchestrates the infusion of local, context-aware style attributes. This hierarchical configuration enables a nuanced control over the balance between style and content, optimizing the fidelity of multi-view 3D generation.
% 后面有时间可以补充一点这两个unet在风格控制和内容保持的实验结果

\begin{figure}[t]
\centerline{ \includegraphics[width=0.7\columnwidth]{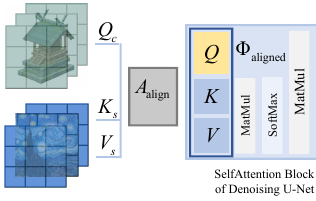}}
\vspace{-5pt}
\caption{\textbf{MultiFusion attention mechanism.} This is designed to align two distinct feature sets to maintain spatial and semantic coherence, by anchoring content-derived query features for geometric consistency across views and infusing style-derived key-value features for high-dimensional texture details.}
\vspace{-12pt}
\label{fig:MFA}
\end{figure}

\noindent\textbf{MultiFussion Attention.} Building upon the observation~\cite{hertz2024stylealignedimagegeneration,chung2024styleinjectiondiffusiontrainingfree,dathathri2020plugplaylanguagemodels,jeong2024visualstylepromptingswapping} that the self-attention within diffusion's latter U-Net architecture acts as a spatial and semantic coherence regulator, our key idea, as depicted in Fig.~\ref{fig:MFA}, is to guide the generation process by aligning two distinct feature sets in diffusion process: (1) Content Features: the query features from the content image, serving as anchors for the geometric consistency across the multiple views. (2) Style Features: key-value features derived from the style image, ensuring that the generated views inherit the desired visual characteristics. Specifically, content-derived query features $\bm{Q}_{c}$ exemplify a probability space over the underlying spatial and geometric feature domain, while the style-derived key-value features $\bm{K}_{s}$ and $\bm{V}_{s}$ encode high-dimensional texture and stylistic nuances. Thus, the spatial coherence and stylistic variation are elegantly orchestrated in the constructed latent space. The relevant concept can be expressed as follows:
%这个是不是有点太扯了
% \textbf{Content-Style Feature Space:}
\begin{equation}
\begin{split}
 \left(\Omega_c, \bm{Q}_{\text{c}}, \mathcal{F}\right) , \quad & \mathcal{F} : \bm{Q}_{\text{c}} \to \Omega_c :\mathbb{R}^n, \\
 \left(\Omega_s, \bm{K}_{{s}}, \bm{V}_{{s}}, \mathcal{G}\right) , \quad & \mathcal{G} : \bm{K}_{{s}}, \bm{V}_{{s}} \to \Omega_s :\mathbb{R}^m, 
\end{split}
\end{equation}
where $\Omega_c$ denotes the spatial domain, encapsulating multi-view geometric invariants, and $\Omega_s$ is the stylistic domain, representing the set over which high-dimensional stylistic attributes, such as texture and pattern variations. 
% \item $\bm{Q}_{\text{c}}$: the Query Features from the content image in latter diffusion process.
% \item $\bm{K}_{\text{s}}, \bm{V}_{\text{s}}$: The Key and Value features derived from the style image, encoding the intricate stylistic nuances required for texture and appearance transfer in the diffusion process.
% \item $\mathcal{F}$: A mapping that projects the characteristics of $\bm{Q}_{\text{c}}$ onto the spatial domain $\Omega_c$, formalizing how latent content features are organized and distributed over $\Omega_c$.
% \item $\mathcal{G}$: A mapping that projects the characteristics of $\bm{K}_{\text{s}}$ and $\bm{V}_{\text{s}}$ onto the stylistic domain $\Omega_s$, formalizing the way style features are structured and distributed over $\Omega_s$.
% \end{itemize} 

Next, we introduce a MultiFusion Attention operator \( \mathcal{A}_{\text{Align}} \) to obtain the fused feature set $\boldsymbol{\Phi}_{\text{aligned}}$ under the condition of the target view sets $\mathcal{M}$:
\begin{equation}
 \boldsymbol{\Phi}_{\text{aligned}} \!= \!\mathcal{A}_{\text{Align}}\left(\left\{\bm{Q}_{c}^{(m)}\right\}_{m \in \mathcal{M}}, \bm{K}_{s}, \bm{V}_{s} \mid \Omega_c, \Omega_s \right),
\end{equation}

% \begin{itemize}
% \item \( \boldsymbol{\Phi}_{\text{aligned}} \): The fused feature set after content and style features are aligned and integrated.
 % \item \( \mathcal{A}_{\text{Align}} \): an operator denoting the process of aligning features with predefined weights.
 % \item \( \mathcal{M}, m \): the set of all target views and each view in it.
% \item \( \mid \Omega_c, \Omega_s\): The relevant condition under the spatial and stylistic domains. 
% \end{itemize}

Following the principles of the basic Attention Theory~\cite{vaswani2023attentionneed}, in a specific diffusion step, the operator \( \mathcal{A}_{\text{Align}} \) can be defined as:
\begin{equation}
\small
 \mathcal{A}_{\text{Align}}^{(t)} \!= \!\operatorname{softmax}\frac{\lambda_t \cdot \bm{\beta} \cdot \left( \bm{Q}_c^{(t)} + \bm{Q}_c^{\text{p}} \right) \cdot \left(\bm{K}_s^{(t)}\right)^\top}{\sqrt{d}} \cdot \bm{V}_s^{(t)},
\end{equation}
where $\lambda_t$ is the time step, which is used to control the influence of the style transfer at different diffusion steps, and $\bm{\beta}=\left(\beta_{c}, \beta_{p}\right)^\top, \beta_{c}+\beta_{p}=1$ denotes the blending coefficient for query preservation, which controls how much of the content's original query features are retained.

\subsection{Sparse-View Spatial Reconstruction}
\label{sec:stage2}
\begin{figure}[t]
\centerline{ \includegraphics[width=\columnwidth]{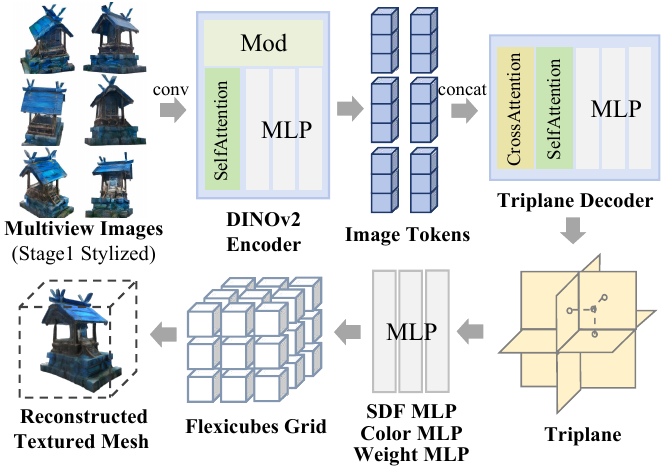}}
\vspace{-5pt}
\caption{\textbf{The procedure of the stylized 3D Reconstruction.} Multi-view images generated in the first stage are used to reconstruct a high-quality 3D object, leveraging SDF-based implicit representations to accurately encode 3D geometry and ensure smooth, flexible surface definitions.}
\vspace{-15pt}
\label{fig:recons}
\end{figure}

In the second stage of Style3D, we focus on Sparse-view Spatial Reconstruction as described in Fig.~\ref{fig:recons}, where the multi-view images generated in the first stage are used to reconstruct a high-quality 3D object. This stage builds upon SDF-based implicit representations~\cite{park2019deepsdflearningcontinuoussigned} for geometry reconstruction. As demonstrated in previous works, SDF representations allow for precise encoding of the 3D geometry, where the values within the field are positive inside the object and negative outside, providing a smooth and flexible way to define 3D surfaces.
% involving encoding the features from the multi-view images and decoding them into a 3D mesh that accurately represents the object

\noindent\textbf{Feature Encoding.} The generated multi-view images in the former stage have integrated content and style features. Next, these multi-view images are processed through a feature encoder to extract compact representations. Our approach builds upon transformer-based architectures, similar to works like OpenLRM~\cite{openlrm}, which uses a pre-trained Vision Transformer (ViT)~\cite{dosovitskiy2021imageworth16x16words} for encoding image features. The ViT model, trained on self-supervised tasks like DINO~\cite{Zhang2022DINODW}, excels at capturing both global structural information and detailed texture patterns in images. The encoding process results in a unified latent feature space that elegantly encapsulates both the object’s geometric structure and the desired stylistic texture. This enables the efficient extraction of both geometric and stylistic features, which are essential for generating high-quality, stylized 3D reconstructions.
%The encoder captures not only the geometric structure but also the stylistic properties. 
%extracts compact, multi-dimensional representations of the object. The encoder captures not only the content’s geometric structure but also the stylistic properties from the style image, creating a unified latent feature space that encapsulates both the geometry and texture of the 3D object.
%These multi-view images are then passed through a feature encoder, which extracts compact representations that encode both the geometric structure from the content image and the stylistic features from the style image. The encoding process results in a unified latent feature space that elegantly encapsulates both the object’s geometric structure and the desired stylistic texture.

\noindent\textbf{Triplane Decoding.}
Once the content and style features are encoded into the unified latent space, the next critical phase is Triplane Decoding. In this step, the multi-view features are projected onto three orthogonal planes (X-Y, X-Z, Y-Z) by the triplane decoder. Each of these planes captures a distinct spatial dimension, enabling effective reconstruction of the 3D structure. The triplane representation naturally captures the 3D geometry, similar to pixel-based methods like NeRF~\cite{mildenhall2020nerfrepresentingscenesneural}, but with the advantage of accommodating multi-view integration. 

By aligning content-derived geometric features and style-specific texture features in the latent space, the decoder integrates spatial coherence with stylistic fidelity. This multi-view-aligned representation ensures that the 3D object is consistent across different viewpoints, with both the geometric structure and stylistic attributes preserved. Importantly, the method eliminates the need for a global coordinate system or predefined camera poses, making it flexible and adaptable to arbitrary viewpoint configurations. This flexibility allows for precise control over the visual style while maintaining consistency across different views. The triplane decoder thus ensures both geometric accuracy and style coherence in the generated 3D object.
%The decoder synthesizes these multi-view inputs and projects them onto the triplane, facilitating an efficient and accurate representation of the object’s 3D structure. The resulting output is a 3D mesh that preserves the object's geometric layout while inheriting the desired visual style, all while maintaining consistency across multiple views.

\noindent\textbf{FlexiCubes Integration.}
In our approach, we utilize the FlexiCubes~\cite{10.1145/3592430} framework for 3D mesh extraction, which plays a crucial role in accurately reconstructing the 3D object’s surface from implicit features. By leveraging FlexiCubes, we effectively navigate the complexities of implicit 3D representations and explicit mesh extraction, allowing for high-quality surface reconstruction directly from triplane implicit fields. This method not only ensures geometric accuracy but also offers flexibility to adapt the mesh to stylistic transformation from the multi-view stylization stage. Its advantage lies in its ability to capture high-fidelity geometric details without relying on traditional methods like Marching Cubes~\cite{10.1145/37402.37422}, which often struggle with topological ambiguities and fail to maintain precise geometric accuracy. The integration of SDF-based implicit fields with explicit mesh extraction ensures that we can represent detailed 3D geometry, while preserving the complex style transfer that was applied during the multi-view generation stage.

To enhance mesh reconstruction, we incorporate additional MLPs (Multi-Layer Perceptrons) to predict the deformation and weights required for FlexiCubes. This enables flexible surface transformations, allowing the 3D mesh to adapt to the style and content features while maintaining consistency with the stylistic variations derived from the style image. By combining SDF-based implicit fields for geometry with explicit mesh extraction through FlexiCubes, we ensure that the 3D object captures both geometric fidelity and style consistency. This dual approach enhances the precision and flexibility of the reconstruction, achieving a balanced integration of geometric accuracy and stylistic coherence, and making the method well-suited for large-scale, style-consistent 3D generation tasks.

%We leverag the FlexiCubes framework, which extracts high-fidelity 3D mesh surfaces from the triplane implicit fields. By using FlexiCubes, we are able to optimize mesh geometry directly within the implicit space, ensuring that the 3D surface is not only geometrically accurate but also adaptable to the style transformation introduced in the first stage.
%In the second phase of Style3D, we further enhance the reconstruction process by incorporating a mesh representation for efficient training. This stage

%To further enhance the mesh reconstruction, we incorporate additional MLPs (Multi-Layer Perceptrons) into the framework to predict the deformation and the weights required for FlexiCubes. This allows us to achieve flexible surface transformations, enabling the 3D mesh to undergo the necessary changes dictated by the style and content features. The deformation and weighting process ensures that the 3D surface adapts smoothly and maintains consistency with the stylistic variations derived from the style image.

%By combining SDF-based implicit fields for geometry and explicit mesh extraction via FlexiCubes, we ensure that the 3D object not only captures the desired geometric fidelity but also integrates style consistency from the input style images, achieving a balanced integration of geometric accuracy and stylistic coherence. This dual approach enhances the precision and flexibility of the 3D reconstruction, making the final output more faithful to both the content and style, and well-suited for large-scale, style-consistent 3D generation tasks.

\noindent\textbf{Geometric Supervision and Regularization.}
To further enhance the quality and stability of the reconstruction, we incorporate geometric supervision using depth and normal maps same as ~\cite{xu2024instantmesh}, which ensure that the 3D mesh aligns with the real-world geometry. In addition, regularization techniques are applied to prevent overfitting and to maintain the stability and consistency of the learned 3D mesh throughout the reconstruction process. Specifically, depth supervision encourages accurate depth predictions by aligning the rendered depth maps with the ground truth, while normal supervision ensures that the normals of the generated mesh closely match the ground truth normals, preserving essential geometric details. Furthermore, regularization is applied to the FlexiCubes weights and deformations, guiding the transformation of the mesh in a smooth and realistic manner. Together, these mechanisms help to achieve a balanced integration of geometric fidelity and stylistic coherence, ensuring that the 3D object remains stable and consistent while adapting to both the content and style features.

%This multi-stage training strategy, combining implicit SDF representations and explicit mesh-based optimization, 

%The loss function for this stage combines the following components:Depth Loss: Enforcing accurate depth predictions between the rendered and ground truth depth maps.Normal Loss: Ensuring that the normals of the generated mesh match the ground truth normals, preserving geometric details.Regularization Loss: Applied to the FlexiCubes weights and deformations to ensure smooth and realistic mesh transformation.
%This multi-stage training strategy, combining implicit SDF representations and explicit mesh-based optimization, enables Style3D to achieve high-quality 3D meshes that integrate both stylistic fidelity and geometric accuracy, all while leveraging minimal input data.

\section{Experiments}

\subsection{Experimental Settings}

\textbf{Pre-trained Models.} For stylized multi-view generation, we directly use the pre-trained Zero123++~\cite{shi2023zero123singleimageconsistent} model. This model was trained on the Objaverse dataset~\cite{deitke2023objaverse}, consisting of 260K filtered 3D shapes with low-quality instances removed. During the training process in Zero123++, each object was rendered to 32 random viewpoints sized at $512\times512$ pixels, and a subset of 6 images was used as input, with 4 additional images for supervision resized to $320\times320$ pixels.
%The pre-trained model uses a Vision Transformer (ViT) for feature encoding, trained on self-supervised tasks like DINO~\cite{}.

\noindent\textbf{Evaluation Dataset.} 
For content image, we use three publicly available datasets: Google Scanned Objects (GSO)~\cite{downs2022googlescannedobjectshighquality}, OmniObject3D (Omni3D)~\cite{wu2023omniobject3d} and NeRF-synthetic~\cite{mildenhall2020nerfrepresentingscenesneural}. We select 100 objects separately from the GSO dataset and the Omni3D dataset. Images are rendered along a uniform orbiting trajectory, with elevations set at {30°, 0°, -30°}, similar to InstantMesh.
For style images, we utilize the public dataset WikiArt~\cite{artgan2018}, which includes over 50k images from 195 different artists. We randomly pick 100 images, similar to StyleRF~\cite{liu2023stylerfzeroshot3dstyle} and StyleGaussian~\cite{liu2024stylegaussianinstant3Dstyle}, while without retraining on it.
%The WikiArt dataset is a collection of artworks, featuring works. The dataset showcases a diverse range of artworks, with varying degrees of realism and stylization, and spans the entire history of art, from cave paintings to modern private collections. Besides, we also add a bruch of a text-to-image Stable Diffusion to provide scalability of stylization from text prompt.

%From the GSO dataset, 100 objects are selected, while for Omni3D, we choose objects from 28 categories, totaling 130 objects. Multi-view images are rendered from 21 views in a uniform orbiting trajectory with varying elevations of {30°, 0°, -30°}. For Omni3D, we also use 16 benchmark views randomly sampled from the top semi-sphere of each object.

%\noindent\textbf{Network Architecture.} Style3D network architecture extends the Zero123++ pipeline, which builds upon the 32-layer Stable Diffusion 2 framework. It incorporates a customized dual U-Net architecture. Our approach introduces a MultiFusion Attention mechanism in the latter up blocks of the U-Net's self-attention layers, with two parameters $\lambda$ and $\bm{\beta}$ for content preservation and style injection.

%\subsection{Main Results}
\subsection{Qualitative Evaluation}
Style3D is designed to generate stylized 3D objects from given images without the need for retraining, distinguishing itself from existing methods in this area. Given the scarcity of existing approaches addressing this specific task, we compare Style3D with two state-of-the-art  generation-based methods and two optimization methods.

\noindent\textbf{Comparisons with Generation Methods.}
We further compare Style3D with two state-of-the-art diffusion-based methods, TEXTure~\cite{richardson2023texturetextguidedtexturing3d} and Paint-it~\cite{youwang2024paintittexttotexturesynthesisdeep}. These methods, which also perform 3D stylization, require 3D meshes as input and generate textured meshes as output. To ensure a fair comparison, we compare Style3D with baseline methods in two distinct tasks, texture generation and stylization, and add a text-to-image brunch~\ref{rombach2021highresolution} to adapt their input prompts into our input style images. We rendered images from evaluation dataset as content images, while providing the corresponding meshes as their mesh input. 

\begin{figure}[t]
\centerline{ \includegraphics[width=0.85\linewidth]{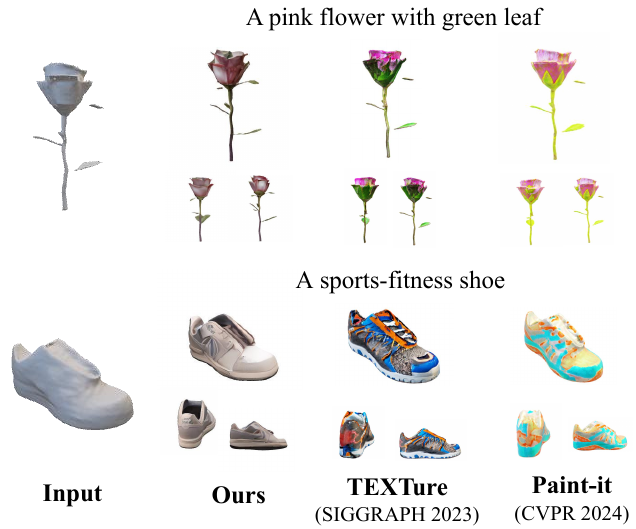}}
\vspace{-6pt}
\caption{\textbf{Qualitative comparison on texture generation.} Style3D is compared with SOTA texture generation methods~\cite{richardson2023texturetextguidedtexturing3d,youwang2024paintittexttotexturesynthesisdeep}. Given the mesh and the corresponding prompt as used in baseline methods, we render a front-view image as the content image and pair it with a image  generated from the prompt as the input. Style3D demonstrates a comparable performance in generating consistent and visually natural 3D textured objects.}
\vspace{-12pt}
\label{fig:comptex}
\end{figure}

For the texture generation task, we used objects and prompts similar to those employed by the baseline methods, rendering the corresponding content and style images. The comparison results, shown in Fig.~\ref{fig:comptex}, reveal that Style3D generates more visually natural results while adhering to the expected texture properties, without the need for a 3D mesh input.

\begin{figure*}[t]
\centerline{ \includegraphics[width=0.96\linewidth]{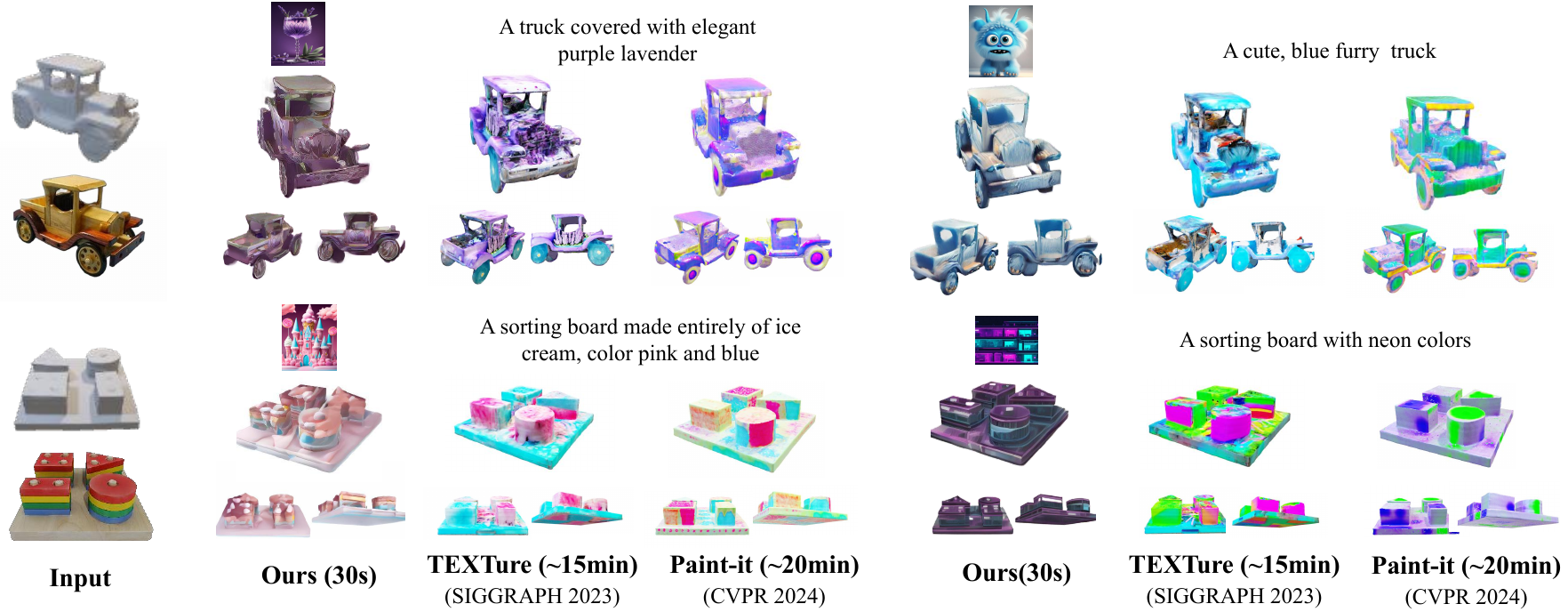}}
\vspace{-5pt}
\caption{\textbf{Qualitative comparison with generation methods on stylization.} The figure illustrates the superior ability of Style3D to adapt and transfer style characteristics, with a content image rendered from the input mesh and a style image generated by the prompt as input, addressing the gaps of existing methods. }
\vspace{-4pt}
\label{fig:compsty}
\end{figure*}

In the style transfer task,  when modifying the prompts to aim for 3D stylization, both TEXTure and Paint-it fail to generate the desired stylized outputs, as shown in Fig.~\ref{fig:compsty}. The baselines tend to produce results that match only the color described in the prompt but fails to capture the full range of stylistic nuances. In contrast, Style3D learns from the style image's features, enabling it to handle a wider variety of style transfer tasks while maintaining the spatial coherence of the object from the content image.

Through both tasks, we demonstrate that Style3D offers superior visual results, addressing limitations that existing methods cannot overcome. Additionally, we observed that baseline methods often generate highly similar outputs for text prompts with similar keywords, even when the meaning differs. Style3D, however, avoids this limitation, generating more varied results. More results can be found in the supplementary materials.

\begin{figure}[t]
\centerline{ \includegraphics[width=\linewidth]{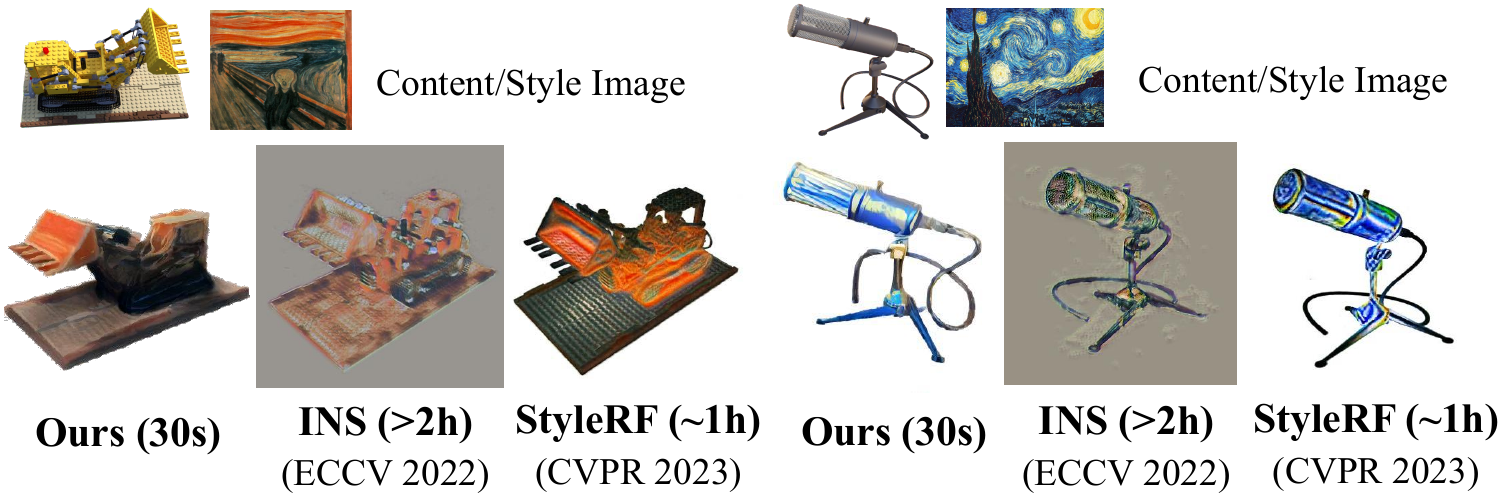}}
\vspace{-3pt}
\caption{\textbf{Comparisons with optimization-based methods.} We compare Style3D with two optimization-based 3D style transfer methods: INS~\cite{fan2022unified} and StyleRF~\cite{liu2023stylerfzeroshot3dstyle}. These approaches typically require style-specific training, making them slow and inefficient for a new style.}
\vspace{-10pt}
\label{fig:compnerf}
\end{figure}

\noindent\textbf{Comparisons with Optimization Methods.}
We evaluate the performance of Style3D with INS~\cite{fan2022unified} and StyleRF~\cite{liu2023stylerfzeroshot3dstyle} on the NeRF-synthetic dataset according to the settings outlined in their respective works. Existing 3D stylization methods typically rely on a large number of views (approximately 100) as input and require training on specific styles or objects. In contrast, Style3D does not require such task-specific training, providing a more flexible and efficient solution for 3D stylization. As shown in Fig.~\ref{fig:compnerf}, the results demonstrate that Style3D not only reduces the input requirements and the stylization time but also successfully transfers stylistic features such as brushstrokes and colors from the style image. This leads to a smoother and more natural stylization effect compared to traditional methods.

\subsection{User Study}
We conducted a user evaluation of five methods by randomly selecting 10 samples from the evaluation dataset. Each user was asked to rate from 1 to 5 for each result on three questions: 1) Whether the result appears visually natural; 2) Whether the texture appearance aligns with the geometric shape; 3) Whether the result matches the given conditions (text or image). 
Evaluations were collected from 30 users, with the criteria conceptualized as Realism (naturalness), Coherence (texture-geometry consistency), and Fidelity (alignment with given conditions). The results are shown as Tab.~\ref{tab:User Study Results}.

\begin{table}[!t]
  \centering
  \small
\caption{\textbf{User study results.} We present Realim, Coherence and Fidelity with SOTA methods. }
  \vspace{-8pt}
  \begin{tabularx}{\linewidth}{@{}l|CCC}
  %$\downarrow$
    \toprule
    Method & Realism $\uparrow$ & Coherence $\uparrow$ & Fidelity $\uparrow$ \\
    \midrule
    INS~\cite{fan2022unified} & 3.32 & 3.71 & 4.23\\
    StyleRF~\cite{liu2023stylerfzeroshot3dstyle} & 3.61 & 3.75 & \textbf{4.30}\\
    TEXTure~\cite{richardson2023texturetextguidedtexturing3d} & 3.82 & 3.91 & 3.86\\
    Paint-it~\cite{youwang2024paintittexttotexturesynthesisdeep} & 3.83 & 4.01 & 4.21\\
    Ours & \textbf{3.89} & \textbf{4.15} &  4.02\\
    \bottomrule
  \end{tabularx}

  \label{tab:User Study Results}
  \vspace{-10pt}
\end{table}

From the user study, it is evident that Style3D excels in realism and coherence, receiving the most favorable feedback. In terms of fidelity, Style3D performs comparably to the other methods. However, it is important to note that the comparison methods, INS~\cite{fan2022unified} and StyleRF~\cite{liu2023stylerfzeroshot3dstyle}, require pre-training on large-scale style datasets and rely on dense viewpoint inputs (100 views) for each object, whereas our method is a single-view, training-free approach. In contrast, Texture and Paint-It do not perform 3D reconstruction as part of their tasks, instead focusing solely on texture generation for input 3D meshes. While Paint-It performs well across all dimensions, it requires approximately 20 minutes of training while ours require only 30s.

\subsection{Ablation Study}

\begin{figure*}[t]
\centerline{ \includegraphics[width=0.94\textwidth]{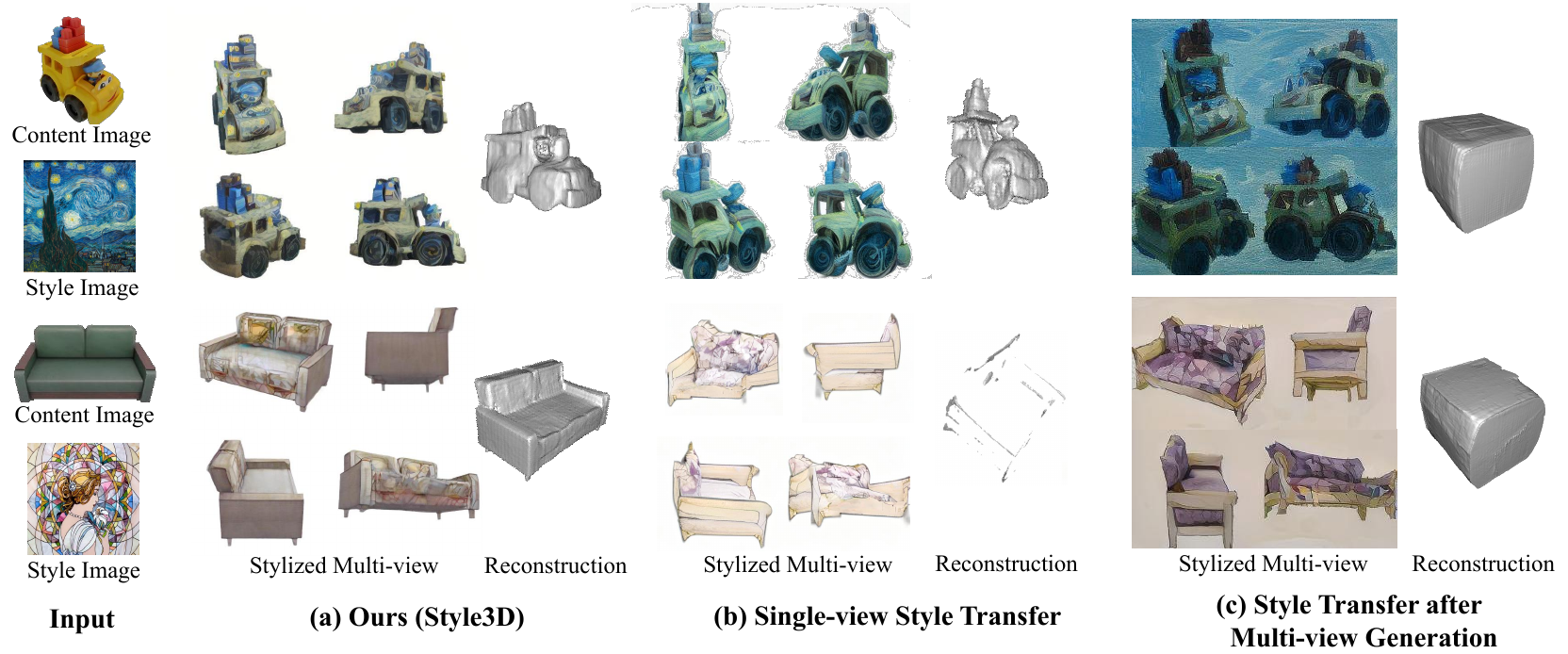}}
\vspace{-5pt}
\caption{\textbf{Ablation study on MultiFusion attention mechanism.} We compare our Style3D (a) with two potential methods for stylized 3D object generation: generating multi-view images from a stylized content image (b), applying style transfer after the generation of multi-view images (c).}
\vspace{-10pt}
\label{fig:ablation}
\end{figure*}

\noindent\textbf{MultiFusion Attention Mechanism.} To validate the effectiveness of the MultiFusion attention mechanism, we conducted a series of ablation experiments, as shown in Fig.~\ref{fig:ablation}, to examine its impact by separating the content and style transfer processes. (1) First applying a state-of-the-art 2D attention-based style transfer method~\cite{chung2024styleinjectiondiffusiontrainingfree} to stylize the content image, and then feed the stylized single-view image into an advanced 3D generation model~\cite{xu2024instantmesh} for image-to-3D conversion. (2) Apply style transfer individually to each of the views directly generated from content image, and then perform 3D reconstruction. 

%请注意我们的方法具有较好的几何保持的性能，尤其体现在精细的结构和清晰的边缘；然而，另外两个则在多视图中就遭受较大的几何扭曲，导致三维重建乱糟糟或重建失效（重建网络将多视图简单认为是一个立方体的六个侧面）

It is important to note that our method demonstrates superior geometric preservation, particularly evident in the fine structure and sharp edges. In contrast, the other two schemes suffer from significant geometric distortions across multiple views, resulting in poor or failed 3D reconstructions. In some cases, the reconstruction networks even treat the multi-view inputs as simply the six faces of a cube, leading to ineffectiveness.

%将产生不利于重建的结果
Through these experiments, we demonstrate that the separation of content and style processing, either sequentially or individually per view, leads to a breakdown in the preservation of both geometry and style. In contrast, MultiFusion Attention ensures that the interaction between content and style features is harmoniously aligned across multiple views, resulting in 3D stylized objects that maintain both geometric fidelity and stylistic coherence.

\begin{figure}[t]
\centerline{ \includegraphics[width=0.95\linewidth]{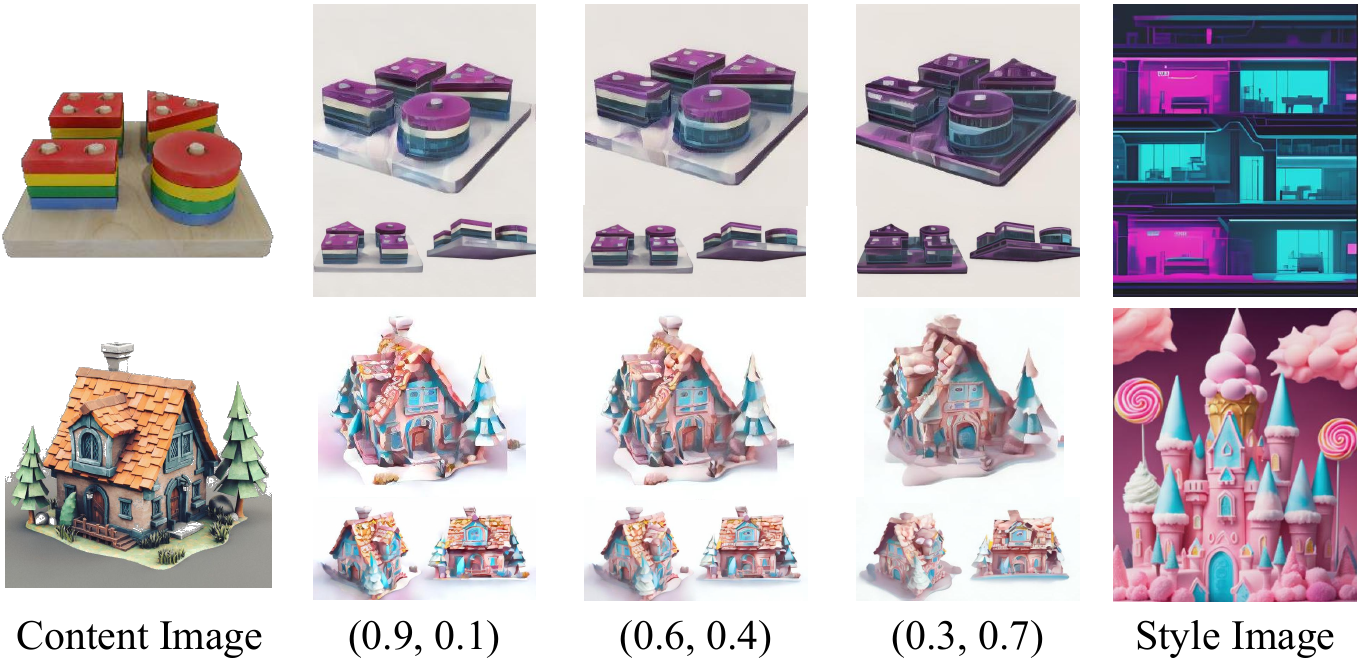}}
\vspace{-7pt}
\caption{\textbf{Visual results of different $(\beta_c,\beta_p)$.} Different values of $\bm{\beta}$ influence the degree of stylization, and the results can be adjusted by tuning this parameter. }
\vspace{-5pt}
\label{fig:beta}
\end{figure}

\begin{figure}[t]
\centerline{ \includegraphics[width=0.95\columnwidth]{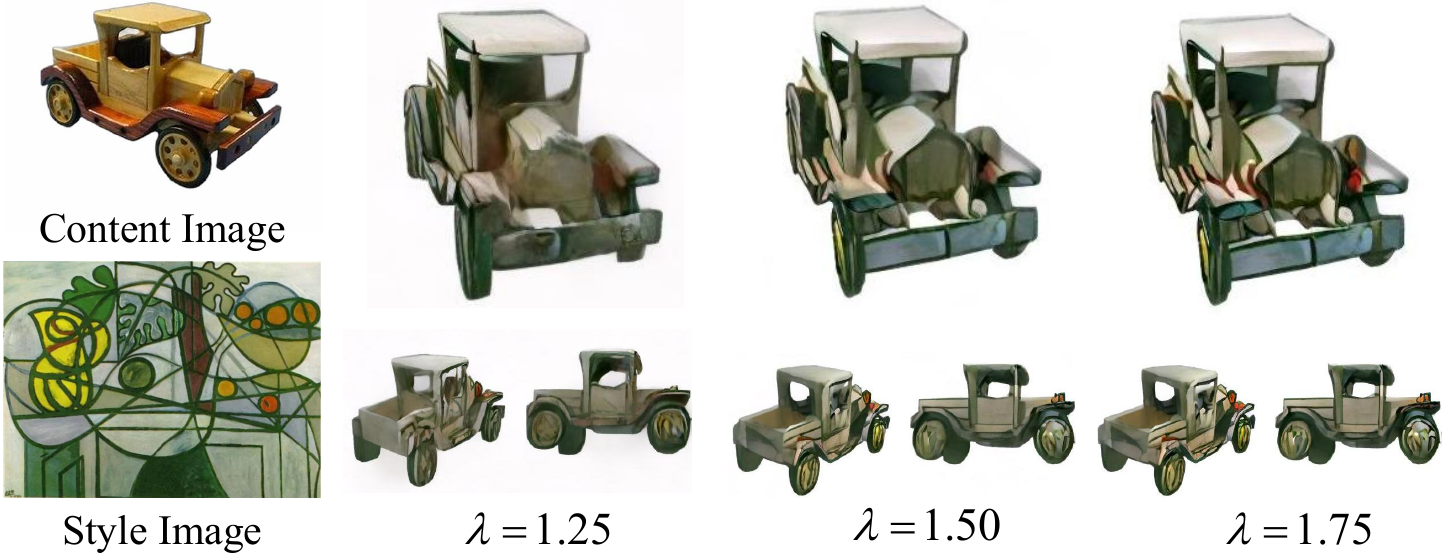}}
\vspace{-5pt}
\caption{\textbf{Visual results of different $\lambda$.} As the value of $\lambda$ increases, the degree of stylization in the results improves, allowing for the display of more detailed features.}
\vspace{-12pt}
\label{fig:lambda}
\end{figure}

\noindent\textbf{Visual Results of different controllable parameters.}
Additional visual results are presented in Fig.~\ref{fig:beta} and Fig.~\ref{fig:lambda}, showcasing the impact of varying parameters within our MultiFusion Attention mechanism. By adjusting the content preservation and style injection factors, we demonstrate the flexibility and control offered by Style3D in achieving the desired balance between geometric fidelity and stylistic transformation. 

\section{Conclusion}
In this work, we introduced Style3D, a novel framework for generating stylized 3D objects from content and style images. By decoupling the process into multi-view dual-feature alignment and sparse-view spatial reconstruction, we effectively integrate geometric consistency and stylistic fidelity without requiring task-specific training. The key innovation, MultiFusion Attention, ensures the preservation of spatial coherence while transferring the style across multiple views. Our approach outperforms traditional methods by offering a scalable, efficient, and high-quality solution for 3D stylization. Extensive experiments have validated that Style3D achieves superior performance, demonstrating its potential for large-scale 3D asset generation.

{
    \small
    \bibliographystyle{ieeenat_fullname}
    \bibliography{reference}
}

\clearpage
\setcounter{page}{1}
\begin{appendix}
\setcounter{table}{0} 
\setcounter{figure}{0}
\setcounter{equation}{0}
\renewcommand\thetable{A.\arabic{table}}
\renewcommand\thefigure{A.\arabic{figure}}
\renewcommand\theequation{A.\arabic{equation}}

\maketitlesupplementary

\section{Implementation Details}
\subsection{Multi-view Generation Network}

Our method is implemented using a single A800 GPU (cuda 12.1, Ubuntu 18.04), extending the zero123++ pipeline, based on the 32-layer Stable Diffusion 2 framework (consisting of 12 down-sampling, 2 middle, and 18 up-sampling layers). The MultiFusion Attention mechanism operates within the latter up-sampling blocks of the U-Net’s self-attention layers and is parameterized by $\lambda$ and $\bm{\beta}$ to control content preservation and style injection. Specifically, we set $\bm{\beta}=(0.4, 0.6)$ and $\lambda = 1.5$, with diffusion step of 65 and a fixed seed of 42. The MultiFusion Attention mechanism is applied to the last five up-sampling block layers as follows (Table~\ref{tab:layer}):

\begin{table}[h]
    \centering
    \caption{\textbf{Attention layers with MultiFusion Attention.} We employ MultiFusion Attention within the last five up-sampling blocks of the U-Net’s self-attention layers.}
    \begin{tabular}{c|l}
        \hline
        \textbf{Block} & \textbf{Attention Layer} \\
        \hline
        up blocks.3 & attentions.0.transformer blocks.0.attn2 \\
        up blocks.3 & attentions.1.transformer blocks.0.attn1 \\
        up blocks.3 & attentions.1.transformer blocks.0.attn2 \\
        up blocks.3 & attentions.2.transformer blocks.0.attn1 \\
        up blocks.3 & attentions.2.transformer blocks.0.attn2 \\
        \hline
    \end{tabular}

    \label{tab:layer}
\end{table}

In the denoising stage of our stylization approach, the content image serves as a conditioning input for zero123++'s primary U-Net. This configuration enables the extraction of \( Q_c^p \) to preserve the structural integrity of the original content.

\subsection{Details in Reconstruction Stage}
To incorporate multi-view inputs, AdaLN camera pose modulation layers are introduced in the ViT image encoder, enabling pose-aware output image tokens. Additionally, the source camera modulation layers in the triplane decoder are removed to simplify the architecture. The NeRF reconstruction losses in this stage combines image loss, perceptual loss, and mask loss, same as those in ~\cite{xu2024instantmesh}:

\begin{equation}
\begin{aligned}
L_1 = \sum_i &\| \hat{I}_i - I_i^{\text{gt}} \|^2 \\
&+ \lambda_{\text{lpips}} \cdot \sum_i L_{\text{lpips}}(\hat{I}_i, I_i^{\text{gt}}) \\
&+ \lambda_{\text{mask}} \cdot \sum_i \| \hat{M}_i - M_i^{\text{gt}} \|^2,
\end{aligned}
\end{equation}

FlexiCubes are integrated into the reconstruction model to extract mesh surfaces from triplane implicit fields. The density MLP of the triplane NeRF renderer is repurposed to predict signed distance functions (SDF), while two additional MLPs predict the deformation and weights required by FlexiCubes. The density-to-SDF conversion is achieved through a initialization for the weight and bias of the last SDF MLP layer to ensure that the SDF field aligns with the desired zero-level set boundary at the start of training. The loss function to predict SDF includes following geometric terms as in ~\cite{li2024instant3d}:

\begin{equation}
\begin{aligned}
L_2 = L_1 
&+ \lambda_{\text{depth}} \sum_i M_i^{\text{gt}} \odot \| \hat{D}_i - D_i^{\text{gt}} \|_1 \\
&+ \lambda_{\text{normal}} \sum_i M_i^{\text{gt}} \odot (1 - \hat{N}_i \cdot N_i^{\text{gt}}) \\
&+ \lambda_{\text{reg}} L_{\text{reg}},
\end{aligned}
\end{equation}

During training, we set 
\(\lambda_{\text{depth}} = 0.5\), \(\lambda_{\text{normal}} = 0.2\), and \(\lambda_{\text{reg}} = 0.01\), 
with a learning rate starting at \(4.0 \times 10^{-5}\) and decaying via cosine annealing.

%重建端的相关参数与细节

%\subsection{User Study Design}
%User Study Design
%To assess the performance of five different methods, we conducted a user study using 10 randomly selected samples from the test set. Video demonstrations were provided to showcase the 3D visual effects of the results. A total of 30 participants evaluated the samples based on three questions: 

\section{Additional Visual and Quantitative Results}

\renewcommand\thetable{B.\arabic{table}}
\renewcommand\thefigure{B.\arabic{figure}}
\renewcommand\theequation{B.\arabic{equation}}

We present additional visual results and a comprehensive evaluation of Style3D through additional qualitative and quantitative comparisons with existing state-of-the-art methods. These experiments assess Style3D’s performance on both texture generation and style-guided generation tasks, highlighting its flexibility and effectiveness.  

\subsection{Additional Visual Results}
\begin{figure*}[t]
\centerline{ \includegraphics[width=0.88\linewidth]{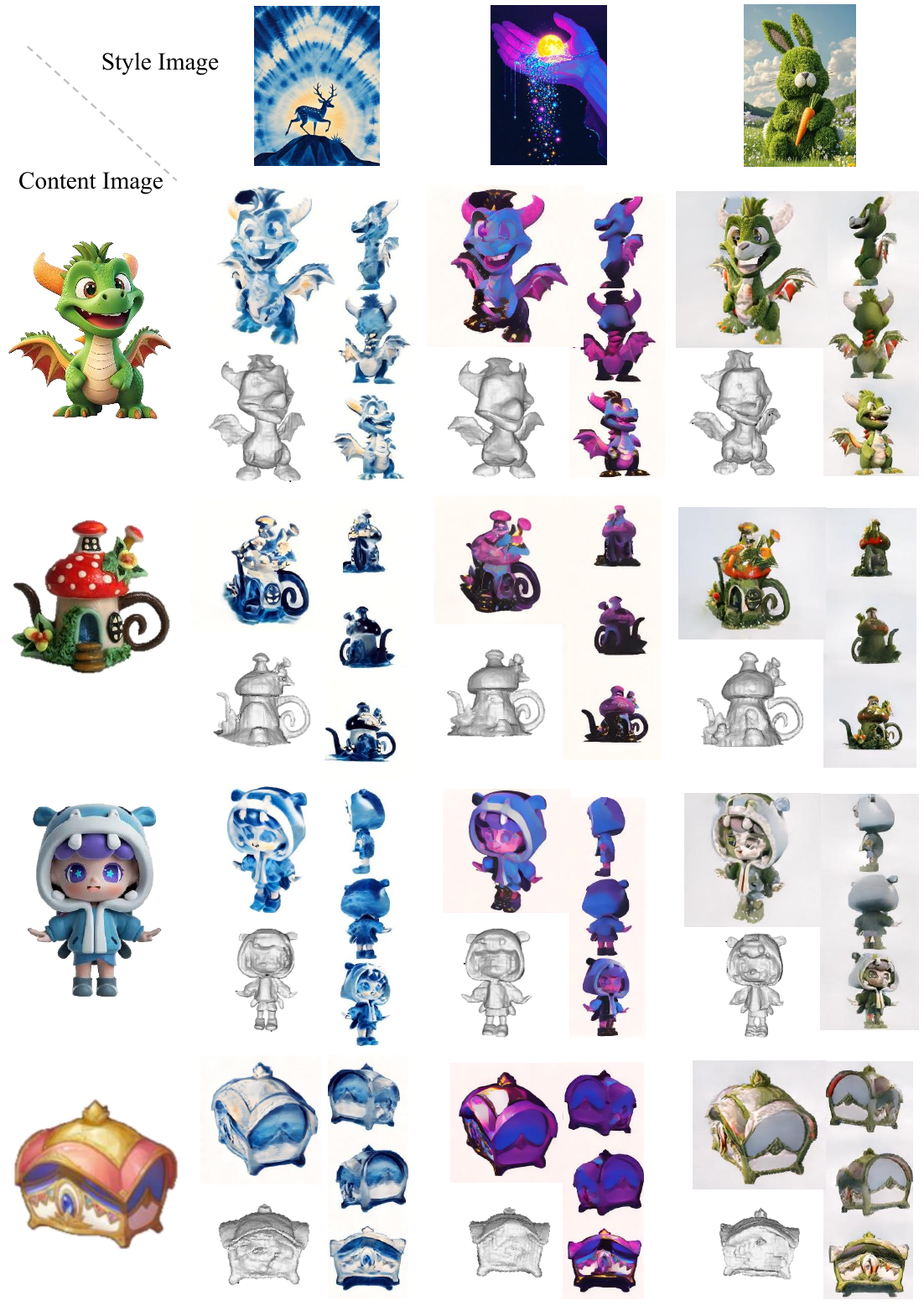}}
% \vspace{-5pt}
\caption{\textbf{Additional Visual results.} Examples of Style3D’s stylized 3D outputs, highlighting its ability to preserve stylistic attributes such as color blending and fine textures.}
% \vspace{-10pt}
\label{fig:additional}
\end{figure*}

Figure~\ref{fig:additional} presents the visual outputs of Style3D, using different combinations of content and style images as input. The results are displayed in both mesh and multi-view output formats. The first column illustrates the application of smooth color gradients and subtle blending effects from the style image. In the second column, oil painting-like textures with luminous qualities are reflected in the outputs. The third column highlights the transfer of plant-inspired fluffy textures, showcasing intricate details. These examples emphasize our ability to faithfully capture and transfer stylistic nuances while preserving the original geometry of the content images.

\begin{figure}[t]
\centerline{ \includegraphics[width=\linewidth]{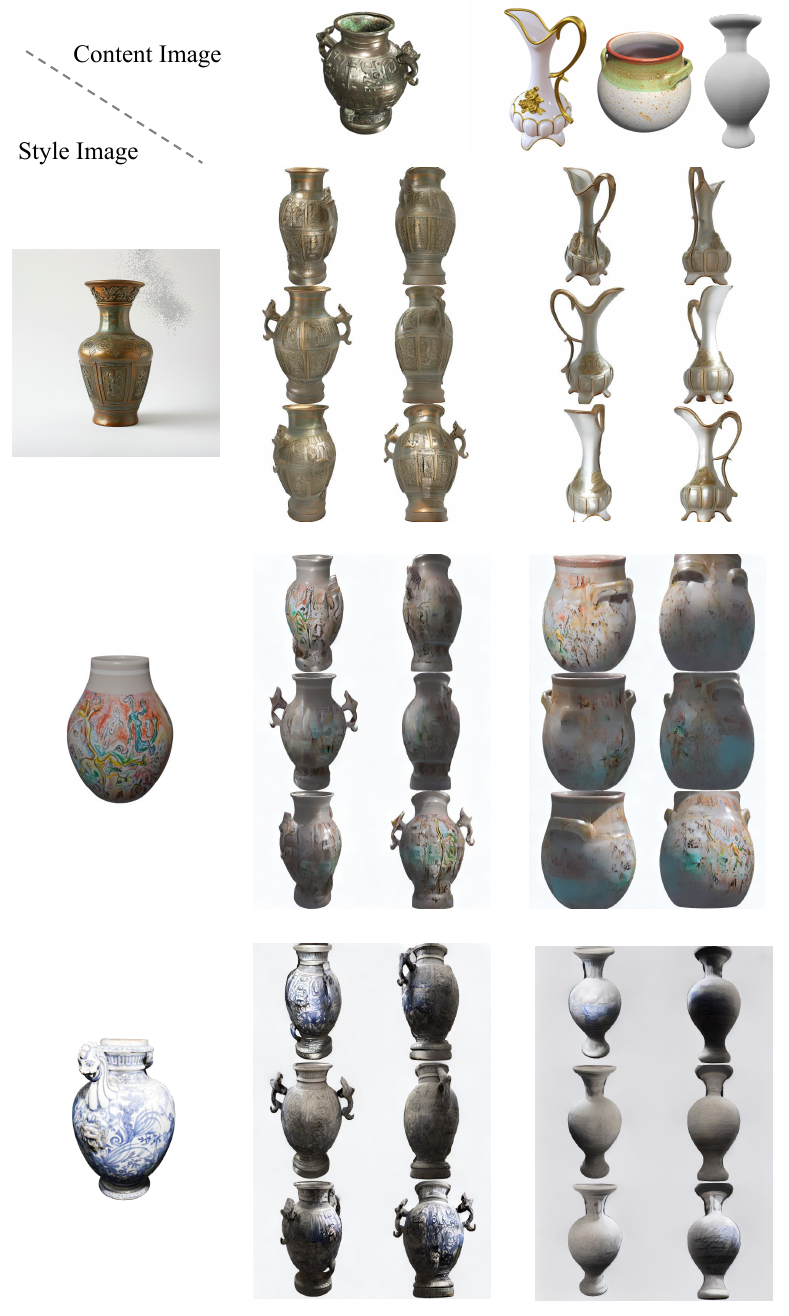}}
% \vspace{-5pt}
\caption{\textbf{Analysis of Style3D’s Adaptability.} Analysis of Style3D’s results under varying content and style complexity, showcasing its robustness and identifying edge cases.}
% \vspace{-10pt}
\label{fig:analysis}
\end{figure}

In Figure~\ref{fig:analysis}, we examine the adaptability of Style3D. Style3D can integrate stylistic elements such as PBR-inspired textures, color gradients, and intricate patterns from style images, demonstrating robustness in transferring these features while maintaining the geometric coherence of content images. Notably, even when the content image contains simple or sparse textures, Style3D can accommodate stylistic changes, especially shown in the second case of the first column. However, challenges arise when the style image exhibits exceptionally detailed or highly complex features, as illustrated in the last row. This indicates that while the method balances 3D consistency and style fidelity, certain scenarios involving extreme stylistic intricacies may require further refinement.

\subsection{Additional Qualitative Comparisons}
\begin{figure}[t]
\centerline{ \includegraphics[width=\linewidth]{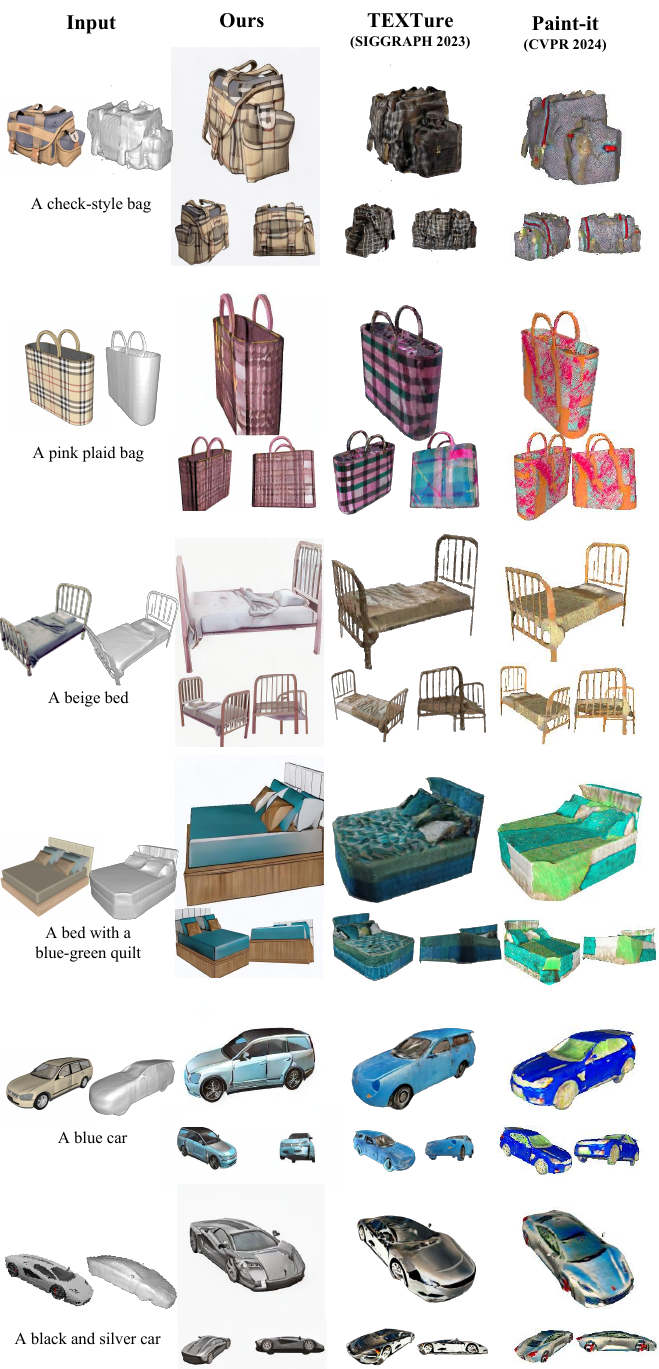}}
% \vspace{-5pt}
\caption{\textbf{Comparison on Texture Generation Task.} Visual comparison between Style3D and baseline methods for texture generation, demonstrating Style3D’s superior style adaptation and consistency.}
% \vspace{-10pt}
\label{fig:suptext}
\end{figure}

In texture generation tasks, as shown in Figure~\ref{fig:suptext}, we provide meshes and textual descriptions as inputs to baseline methods (TEXTure~\cite{richardson2023texturetextguidedtexturing3d} and Paint-it~\cite{youwang2024paintittexttotexturesynthesisdeep}), while Style3D generates results from single-view content and style images without additional training. The results demonstrate that Style3D achieves natural and seamless style transfer, blending content and style features coherently. In contrast, baseline methods often produce visually inconsistent outputs, with noticeable artifacts and limited diversity across multiple views.

\begin{figure*}[t]
\centerline{ \includegraphics[width=0.87\linewidth]{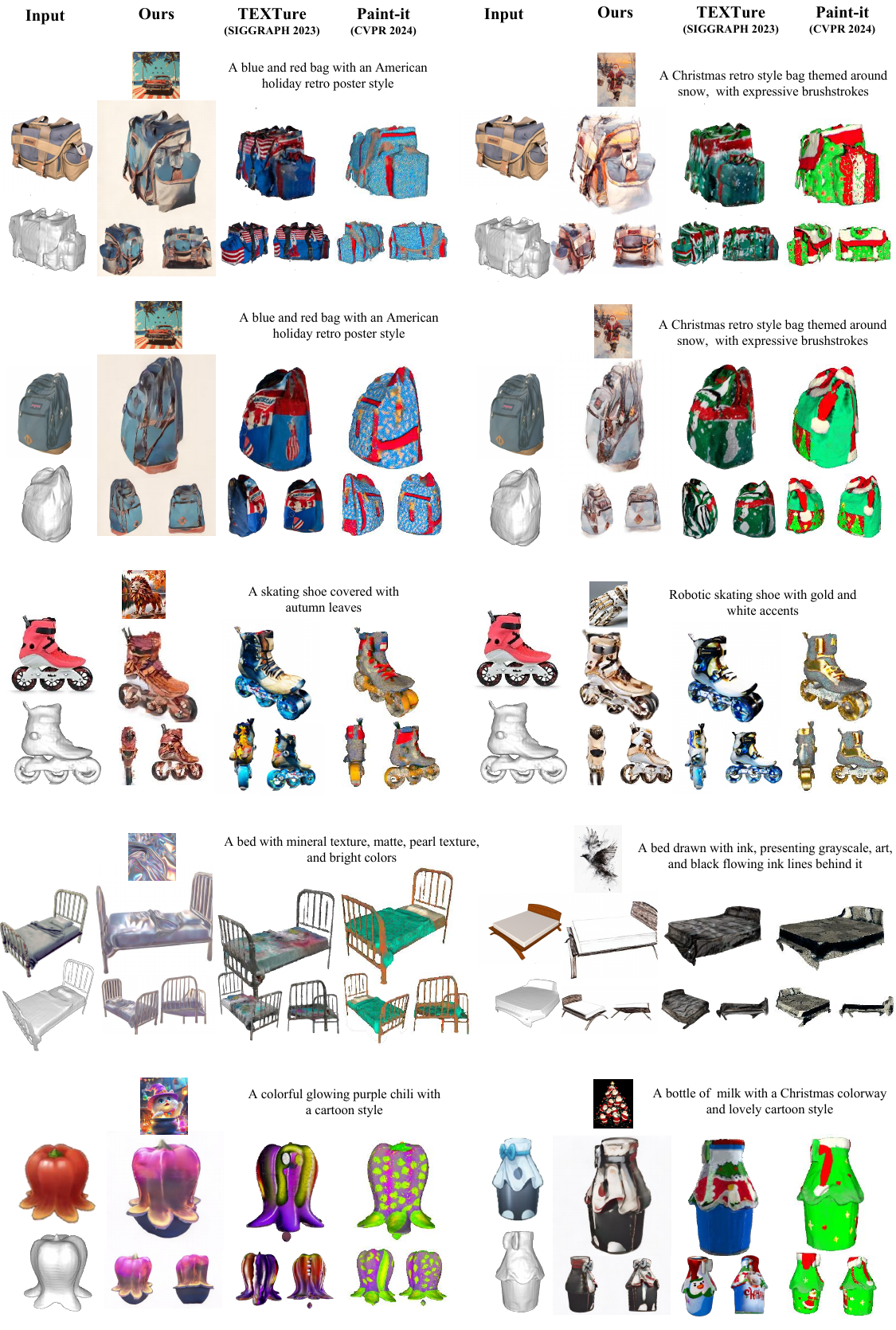}}
\vspace{-15pt}
\caption{\textbf{Comparison on Style-Guided Generation Tasks.} Comparison of Style3D and existing methods under style-guided generation tasks, illustrating Style3D's advantages in spatial coherence and stylistic fidelity.}
% \vspace{-10pt}
\label{fig:supsty}
\end{figure*}

For style-guided generation tasks, Figure~\ref{fig:supsty} illustrates the distinct advantages of Style3D. While baseline methods rely on textual prompts and meshes, their outputs frequently exhibit over-smoothing, lack of diversity, or inconsistent multi-view results. Notably, the Christmas Style examples in the first and last rows show limited adaptability in baseline methods, exhibiting a high degree of similarity for inputs with comparable features. In contrast, Style3D maintains spatial coherence and achieves balanced style transfer, as evidenced by the detailed textures and intricate style adaptations in outputs such as the leaf-covered skating shoe and the pearl-textured bed.

\subsection{Additional Quantitative Comparisons}
To evaluate the effectiveness of our method, we employ two metrics for quantitative analysis: (1) Text-Image CLIP Score, which measures the alignment between the generated image and the target textual description, and (2) Image-Image CLIP Score, which assesses the fidelity of the stylized image to the original content image.

For the evaluation, we select 20 diverse style images to ensure broad coverage of stylistic variations. For 3D scene stylization methods, such as INS~\cite{fan2022unified} and StyleRF~\cite{liu2023stylerfzeroshot3dstyle}, we use scenes from the NeRF synthetic dataset as content images, selecting six random viewpoints for each scene. Similarly, for texture generation methods, including TEXTure~\cite{richardson2023texturetextguidedtexturing3d} and Paint-it~\cite{youwang2024paintittexttotexturesynthesisdeep}, we render textured meshes from six random viewpoints to facilitate a consistent basis for comparison.

\begin{table}[h]
    \centering
    \caption{\textbf{Quantitative Evaluation with CLIP Scores.} Quantitative comparison of Style3D with baseline methods using CLIP metrics, illustrating Style3D's superiority in content fidelity and style diversity.}
    \begin{tabular}{l|cc}
        \hline
        \textbf{Method} & \textbf{Image-Text}$\uparrow$ & \textbf{Image-Image}$\uparrow$ \\
        \hline
        INS~\cite{fan2022unified} & 0.216 & 0.785 \\
        StyleRF~\cite{liu2023stylerfzeroshot3dstyle} & 0.220 & 0.773 \\
        TEXTure~\cite{richardson2023texturetextguidedtexturing3d} & 0.232 & 0.846 \\
        Paint-it~\cite{youwang2024paintittexttotexturesynthesisdeep} & 0.229 & 0.807 \\
        Ours & \textbf{0.241} & \textbf{0.861} \\
        \hline
    \end{tabular}
    \label{tab:clip}
\end{table}

As detailed in Table~\ref{tab:clip}, Style3D demonstrates superior alignment between content and style, outperforming baseline methods.

\section{Limitations}
While Style3D demonstrates significant strengths, it also has certain limitations. The sparse-view reconstruction framework confines its applicability to object-level stylization, limiting its use for full scene reconstruction. Moreover, the trade-off between 3D consistency and style fidelity results in slightly weaker style transfer performance compared to specialized 2D stylization methods. Additionally, The quality of mesh reconstruction could be further enhanced by incorporating more advanced techniques.

%Our method is limited by the constraints of the pre-trained model, which only supports image-to-six-view generation, making scene reconstruction infeasible. This is a consequence of the sparse-view reconstruction framework we employed. Additionally, the stylization effect may be slightly weaker compared to direct 2D stylization due to the trade-off between 3D consistency and stylistic fidelity. Currently, our method does not support direct text-based stylization, though it can be extended using a text-to-image (T2I) branch to provide this capability.

\section{Future Work}
Looking ahead, several avenues exist to extend the capabilities of Style3D. One key direction involves developing fine-grained style control mechanisms that allow for precise manipulation of stylistic elements, enabling users to selectively emphasize specific aspects of the style during the transfer process. Additionally, enhancing the 3D reconstruction quality through the integration of advanced mesh processing techniques and more robust geometric modeling frameworks could significantly improve the fidelity and versatility of the generated results. Furthermore, addressing the current reliance on an auxiliary text-to-image module for text-driven stylization by incorporating direct text-to-style capabilities within the framework would broaden its applicability, particularly for creative design workflows.

\end{appendix}

\end{document}